\documentclass[preprint,12pt]{elsarticle}

\usepackage{hyperref}
\usepackage{graphicx, adjustbox}
\usepackage{amssymb}
\usepackage{amsmath}
\usepackage{wrapfig}

\newcommand{\bottomhline}{\hline \rule{0mm}{-10.0mm}}

\newcommand{\definecaption}[2]{%
   \expandafter\newcommand\csname 4gamma@caption@#1\endcsname{#2}}
\newcommand{\usecaption}[1]{%
   \caption{\csname 4gamma@caption@#1\endcsname\label{#1}}
   }

\definecaption{cotic_layer}{Scheme of the forward pass for the continuous-time convolutional layer used in the COTIC model. We present output at a single point in time $t_3$.}
\definecaption{cotic}{\textbf{COTIC Architecture}: The diagram on the left (Encoder) illustrates the inner layers of the COTIC model, which utilize the states at the timestamps of the original sequence only. The diagram on the right (Final layer) shows the final layer responsible for obtaining the intensity at every point in time.}
\definecaption{sense}{Sensitivity study results. We vary the number of layers, the kernel size, and the activation function (from left to right) to obtain MAE values for the return  time prediction on the Retweet dataset.}
\definecaption{intens}{A graph of the intensity versus time for the artificial data. The blue color indicates the true Hawkes intensity. The orange color indicates the intensity restored via the COTIC model.}

\newcommand{\vecK}{\boldsymbol{k}} 
\newcommand{\vecM}{\mathbf{m}}

\newcommand{\vecY}{\mathbf{y}}

\newcommand{\vecT}{\boldsymbol{\theta}}

\journal{Information Sciences}

\begin{document}

\begin{frontmatter}



\title{COTIC: Embracing Non-uniformity in Event
Sequence Data via Multilayer Continuous
Convolution}


\author[skoltech]{Vladislav Zhuzhel\footnote{Corresponding author, e-mail: vladislav.zhuzhel@skoltech.ru}} 
\author[skoltech]{Vsevolod Grabar}
\author[skoltech]{Galina Boeva}
\author[skoltech]{Artem Zabolotnyi}
\author[skoltech]{Alexander Stepikin}
\author[skoltech]{Vladimir Zholobov}
\author[sber]{Maria Ivanova}
\author[sber]{Mikhail Orlov}
\author[sber]{Ivan A Kireev}
\author[skoltech]{Evgeny Burnaev}
\author[choco,cdtm]{Rodrigo Rivera-Castro}
\author[skoltech]{Alexey Zaytsev}

\affiliation[skoltech]{organization={Skolkovo Institute of Science and Technology},
            addressline={Bolshoy Boulevard 30,  bld. 1}, 
            city={Moscow},
            postcode={121205}, 
            country={Russia}}
\affiliation[sber]{
            organization={Sber AI Lab},
            addressline={Kutuzovsky Avenue, 32}, 
            city={Moscow},
            postcode={121165}, 
            country={Russia}
}
\affiliation[choco]{
            organization={Choco Communication},
            addressline={Hasenheide 54}, 
            city={Berlin},
            postcode={10967}, 
            country={Germany}
}
\affiliation[cdtm]{
            organization={Center for Digital Technology and Management},
            addressline={Arcisstr. 21}, 
            city={Munich},
            postcode={80333}, 
            country={Germany}
}

\begin{abstract}
Event sequences often emerge in data mining. 
Modeling these sequences presents two main challenges: methodological and computational. Methodologically, event sequences are non-uniform and sparse, making traditional models unsuitable. Computationally, the vast amount of data and the significant length of each sequence necessitate complex and efficient models.
Existing solutions, such as recurrent and transformer neural networks, rely on parametric intensity functions defined at each moment. These functions are either limited in their ability to represent complex event sequences or notably inefficient.

We propose COTIC, a method based on an efficient convolution neural network designed to handle the non-uniform occurrence of events over time. 
Our paper introduces a continuous convolution layer, allowing a model to capture complex dependencies, including, e.g., the self-excitement effect, with little computational expense.

COTIC outperforms existing models in predicting the next event time and type, achieving an average rank of $1.5$ compared to $3.714$ for the nearest competitor. 
Furthermore, COTIC’s ability to produce effective embeddings demonstrates its potential for various downstream tasks. 
Our code is open and available at: \url{https://github.com/VladislavZh/COTIC}.
\end{abstract}




\end{frontmatter}

\begin{table}[!ht]
\centering 
\begin{tabular}{lccccc}
\hline
Method &
\begin{tabular}{c}
Next event time\\
MAE, rank  \\
\end{tabular}
& 
\begin{tabular}{c}
Next event type \\
Accuracy, rank \\
\end{tabular}
&
\begin{tabular}{c}
Mean \\
rank  \\
\end{tabular}
\\
\hline
RMTPP~\cite{rmtpp_paper}       & $5.286$ & $6.714$ & $6$ \\
Neural Hawkes~\cite{mei2017neural}   & $4.143$ & $\underline{3.286}$ & $\underline{3.714}$\\
ODETPP~\cite{chen2021neural}   & $6.571$ & $5.571$ & $6.071$ \\
THP~\cite{zuo2020transformer}  & $7.143$ & $5.714$ & $6.428$\\
THP2SAHP~\cite{zhang2020self}   & $6.571$ & $5.143$ & $5.857$ \\
Attentive NHP~\cite{yang2022transformer}  & $\underline{3.714}$ & $4.714$ & $4.214$\\
WaveNet~\cite{vanwavenet} & $4.143$ & $\underline{3.286}$ & $\underline{3.714}$\\
CCNN~\cite{shi2021continuous}  & $5.857$ & $8.571$ & $7.214$  \\
COTIC (ours) & $\mathbf{1.429}$ & $\mathbf{1.571}$ & $\mathbf{1.500}$\\[0.1mm] \bottomhline
\end{tabular}
\caption{Ranks for methods averaged over eight datasets for the problem of next event time and type predictions. A lower rank means that the method is closer to the top-1 method (with the rank one) for a problem. The best results are highlighted with bold font, and the second-best results are underlined.}
\label{tab:ranks}
\end{table}
\section{Introduction}
\label{intro}
Many real-world processes are event sequences, such as bank transactions~\cite{wei2013effective, bazarova2024universal}, purchases at stores~\cite{lysenko2019temporal}, network traffic~\cite{8737622}, series of messages, social media posts, TV program views, and customer flows. 
Consequently, there is a constant need to model such data. For example, one may be interested in predicting the banking activity of a particular customer or estimating the timing of the next purchase at an online marketplace. 
This knowledge provides new business perspectives, such as improved credit default~\cite{babaev2019rnn} and churn~\cite{berger2019user} prediction. 
For example, timely churn prediction leads to more precise and effective marketing campaigns.

Mostly, the next event in a sequence occurs at a random time after the previous event. 
Therefore, the entire sequence is non-uniform and discrete, as events are only recorded at specific time points. 
Additionally, the label of an event is a random variable that can depend on past events. 
This non-uniformity and discreteness necessitate specific formalism for effective modeling.

To address these challenges, event sequences are often modeled as realizations of temporal point processes (TPP) \cite{shchur2021neural}. The key concept in TPPs is intensity, which represents the expected number of events of each type observed during a small period. In classic examples like homogeneous and non-homogeneous Poisson processes~\cite{lawless1987regression, kingman1992poisson}, the intensity is independent of past events. However, this is not the case for scenarios like pandemic spread or order history. A more realistic model is a self-exciting TPP, such as the Hawkes process~\cite{hawkes1971spectra}, where past events influence the future intensity of the event flow.

The diversity of available data makes event sequence modeling complex and necessitates general models that can handle various peculiarities. Constant advances in deep learning have led researchers to explore neural network architectures for modeling event sequences. LSTM-like architectures were among the first proposed~\cite{mei2017neural, rmtpp_paper}. Another approach combined Neural Jump ODEs with Continuous Normalizing Flows~\cite{chen2021neural}.
These methods were later followed by Transformer-based architectures~\cite{zuo2020transformer, yang2022transformer}. The study by Shi et al.~\cite{shi2021continuous} utilized convolutional neural networks (CNNs) for non-uniform time series. However, it showed sub-optimal performance for modeling event sequences due to the use of a single continuous convolution layer and manual grid allocation for subsequent convolutional layers.

Despite the progress made, these models share an important limitation: they recover the intensity function within a fixed exponential or semi-linear parametric family, using it to predict the next event type or time.

In this work, we advance the modeling of event sequences by proposing a COntinuous-TIme Convolutional neural network model (COTIC). COTIC constructs representations of a temporal point process, predicts its intensity, and addresses other downstream problems in an efficient way. 
Our contributions are as follows:
\begin{itemize}
    \item Our novel method, COTIC, is a deep one-dimensional continuous CNN for processing non-uniform event sequences.
    \item We introduce continuous convolutions, avoiding closed-form parametric assumptions for intensity. So, we can model a wide variety of dependencies between events. 
    \item The method also has self-supervised representation learning capabilities due to its generative nature. Compared COTIC's embeddings with other event sequence methods and MiniRocket on the transactions dataset for age bin prediction. COTIC's embeddings provide superior accuracy.
    \item The COTIC outperforms existing approaches, including Transformers, RNNs, Neural ODEs, and baseline CNNs, in predicting the time and type of the next event. \autoref{tab:ranks} shows COTIC's superiority in both problems.
\end{itemize}

\section{Related work}
\label{sec:related_work}
For a comprehensive overview of the current state of neural network models for modeling temporal point processes, we refer interested readers to a recent review~\cite{shchur2021neural}. 
Here, we provide a detailed discussion of the most relevant papers to contextualize our work.

\paragraph{Temporal Point Processes (TPPs)}
A \emph{temporal} point process involves points that lie on a temporal $\mathbb{R}^1$ axis, each associated with a specific time. In a \emph{marked} temporal point process, each point also has an additional mark or a vector of features, which can be represented as belonging to $\mathbb{R}^d$~\cite{yan2019recent}. In some cases, one may consider events that span a duration, where each event is characterized by both a start and an end time \cite{LAI2024120421}. Event sequences are a particular case of realizations of marked temporal point processes.

Self-exciting point processes are of specific interest because they assume that the future intensity of events depends on historical events. Typically, past events increase the intensity of future events, as seen in cases like retweets in social networks~\cite{rizoiu2017hawkes} or the spread of a virus~\cite{chiang2022hawkes}.

\paragraph{Deep Learning for TPPs}
While classic machine learning has been successful in modeling the complex dynamics of event sequences~\cite{zhou2013learning}, the advent of deep learning has brought about greater adaptability and, consequently, better results~\cite{neuralhawkes_paper}. This improvement is due to the flexibility of deep neural networks, which can handle sequences of arbitrary complexity given a sufficiently large training sample.

Additionally, deep learning enhances representation capabilities. Neural networks can generate representations for each moment in time, each event, or an entire event sequence, enabling the solution of various downstream problems by leveraging these representations. However, the primary downside of deep learning is the higher computational cost required to train these models~\cite{zhuzhel2021cohortney}.

A variety of deep learning architectures have been proposed to model temporal point processes in numerous ways. Historically, the first approaches utilized different recurrent architectures, such as Continuous LSTM~\cite{neuralhawkes_paper} and RMTPP~\cite{rmtpp_paper}. Subsequent adaptations involved transformer architectures~\cite{zuo2020transformer,zhang2020self}.

Convolutional neural networks were directly considered in~\cite{shi2021continuous}, where the authors achieved promising results. However, their architecture comprised only two layers, with the first layer solely responsible for converting non-uniform time to uniform time. Papers on state-space layers~\cite{gue2022fficiently} also highlighted the challenge of non-uniform time but did not provide a solution for an efficient non-uniform time state-space layer.

We also note that TPP-based learning occurs in a self-supervised manner, without the need for additional label information, as event types and occurrence times can be extracted directly from the data. However, direct approaches for handling sequential data in a self-supervised way, such as those based on contrastive~\cite{yue2022ts2vec, babaev2022coles}, non-contrastive~\cite{marusov2023non}, and generative strategies~\cite{kenton2019bert}, are complementary to TPPs.

\paragraph{Deep Learning for Sequential Data}
Within the sequential data modeling community, a lively discussion persists about the most effective architecture among Transformers, Recurrent Neural Networks (RNNs), and Convolutional Neural Networks (CNNs). Here, we aim to highlight significant aspects of this debate that are relevant to our research.

RNN based approaches were the first one that were applied for event sequence modelling \cite{neuralhawkes_paper}. RNNs allow the researcher to introduce the evolution of hidden states between events, thus modelling event sequences naturally. However RNNs are inefficient in handling long term dependencies and cannot be parallelized due to the sequential nature of the forward pass.

Transformers often dominate discussions on sequential data due to their remarkable performance in natural language processing. However, it remains an open question whether Transformers are optimal for tasks like time-series modeling. Some transformer-based architectures have been designed to incorporate time-series-specific characteristics, such as trend and seasonality, as demonstrated by \cite{NEURIPS2021_bcc0d400}. For network events, a graph transformer-based model was applied in \cite{LI2023119596}. Nevertheless, Transformers remain relatively inefficient, with a squared complexity on sequence length. Additionally, some researchers contend that, despite the use of positional encoding, Transformers' inherent permutational invariance could lead to a loss of crucial information \cite{Zeng2022AreTE}.

Furthermore, certain studies have suggested the superiority of Convolutional Neural Networks (CNNs) in addressing specific problems. For instance, the issue of a constrained CNN horizon can be circumvented by implementing continuous convolutions, enabling the processing of non-uniform data, as proposed in~\cite{romero2021ckconv}. Another effective strategy balances computational efficiency and predictive quality using a random kernel-based approach, where random kernels generate a broad spectrum of features that are subsequently fed into a classifier~\cite{rocket}.

Additionally,~\cite{kim2023smpconv} demonstrated the effectiveness of continuous convolution without using neural networks, highlighting its advantages. The continuous convolution operation's benefits in large-scale conditions have also been shown. In~\cite{romero2022flexconv} FlexConv was proposed, an ultra-precise operation that enhances the ability to work with convolutional cores with high throughput and affordable core size at a fixed parameter cost.

It is also worth noting that state-space models can be interpreted as a generalization of convolutional and recurrent architectures, and they can compete with Transformers in natural language processing~\cite{dao2023hungry}.

We believe that a careful examination of the domain of event sequence data is essential to identify the best approach in terms of quality and efficiency.

\paragraph{Research Gap}
The primary challenge in applying convolutional neural networks (CNNs) to model temporal point processes (TPPs) lies in the non-uniformity and discreteness of the sequences. Instead of learning values at discrete points of a convolutional kernel, the kernel should be learned as a whole. Recent results have shown that this approach is feasible for multivariate time series~\cite{romerockconv}.

A similar approach was employed in \cite{shi2021continuous}, but due to the limited model size, it was unable to capture the complex dependency structures in the data. Another study, \cite{ccpp2023}, also utilized continuous convolutions but with only one layer, incorporating a recurrent architecture on top of the resulting embeddings. Additionally, this method did not utilize intensities. Our objective is to develop a fully convolutional architecture that predicts the intensity function while maintaining high accuracy and efficiency in both training and inference. This architecture aims to handle long-term dependencies and address various downstream tasks using the generated embeddings.

\section{Methods}
\label{sec:methods}

\begin{figure}
  \centering
  \includegraphics[width=0.75\linewidth]{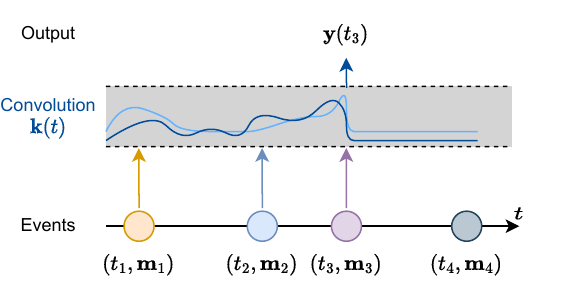}
  \usecaption{cotic_layer}
  \label{fig:cont_conv_layer}
\end{figure}

In this section, we describe our Continuous-Time Convolutions (COTIC) model and highlight its unique features compared to other neural models for processing event sequence data.

\subsection{Preliminaries}
\label{sec:preliminaries}

\paragraph{Notation}
\label{subsubsec:notation}
Suppose we observe \( n \) sequences in a sample \( D = \{\boldsymbol{S}_i\}_{i = 1}^n \). Each sequence $\boldsymbol{S}_i$ consists of tuples \( \boldsymbol{S}_i = \{(t_{ij}, \vecM_{ij})\}_{j = 1}^{T_i} \), where \( t_{ij} \in [0, T_i] \) are event times, such that for \( j > j' \), \( t_{ij} \geq t_{ij'} \), and \( \vecM_{ij} \in \mathbb{R}^d \) denote event type marks for the \( i \)-th sequence in the dataset. If we have a discrete set of possible marks \( \vecM_{ij} \in \{1, \ldots, M\} \), we use an embedding for each mark instead.

We use the notation \( \boldsymbol{S} = \{(t_j, \vecM_j)\}_{j = 1}^{T} \) to refer to an arbitrary sequence from the dataset. For the current event number \( k \), we denote the observation's history as \( \boldsymbol{S}_{1:k} = \{(t_j, \vecM_j)\}_{j = 1}^{k} \) for the first \( k \) events and \( \boldsymbol{S}_{<t} \) for all events with \( t_i < t \).

\paragraph{Intensity Function and Likelihood \cite{rasmussen2018lecture}}
\label{subsubsec:intensity}
When dealing with event sequences, a key challenge is processing the non-uniform nature of time lags between events. One approach is to use the intensity function.

Consider the return time \(\tau_{i} = t_i - t_{i-1}\) and event type \(m\) as random variables. We can introduce their conditional probability density function (PDF) \(f(t, m | \mathcal{H})\), which indicates the probability of an event of the type \(m\) occurring within an infinitesimally small time interval given the history \(\mathcal{H}\) before this interval. Additionally, the conditional survival function \(S(t)\) represents the probability that an event of any type has not occurred by time~\(t\):

\begin{align}
\label{eq:pfd_and_survival}
    &f(t = t_i, m = m_i|\boldsymbol{S}_{1:i - 1}) = \lim_{\Delta t\to +0}{\frac{\mathbb{P}\left(t < t_i\leq t+\Delta t, ~ m=m_i| \boldsymbol{S}_{1:i - 1}\right)}{\Delta t}}, \\
    &S(t) = \mathbb{P}\left(t_i\geq t\right | \boldsymbol{S}_{1:i - 1}) = \int_{t}^{\infty}{f(z| \boldsymbol{S}_{1:i - 1}) dz}.
\end{align}

We denote the conditional intensity function of the sequence as:

\begin{equation}
\label{eq:intensity}
    \lambda_m(t) = \mathbb{E}\left[\frac{dN_m(t)}{dt}\right] = \frac{f(t, m | \boldsymbol{S}_{<t})}{S(t)},
\end{equation}
where \( N_m(t) = \#\{t_j \colon t_j < t \mid m_j = m\} \) is the number of events of type \( m \) that occurred before time \( t \) and \(\boldsymbol{S}_{<t}\) is equivalent to \(\boldsymbol{S}_{1:i - 1}\) and \(t_{i}\geq t\).

The main benefit of using the intensity function instead of the PDF is that it has fewer restrictions. The intensity function must be non-negative and belong to the space of \( L_1 \) functions, meaning its absolute value must be integrable. Unlike the PDF, there is no requirement for the intensity function to integrate to 1.

At the same time, it is straightforward to express the likelihood in terms of the intensity function. For a vector parametric intensity function \(\boldsymbol{\lambda}(t|\vecT) = \{\lambda_m(t|\vecT)\}_{m=1}^{M}\) with the vector of parameters \(\vecT\), the negative log-likelihood for an event sequence \(S_{1:k}\) with no event in the interval \((t_k, T]\) is:

\begin{equation}
\label{eq:likelihood}
    L_{\lambda}(\vecT) = \int_{0}^{T} \sum_{m}\lambda_m(t|\vecT) \, dt - \sum_{j = 1}^{k} \log \lambda_{m_j}(t_j|\vecT).
\end{equation}

Once the intensity \(\boldsymbol{\lambda}(t|\vecT)\) is defined, the model can be optimized via the likelihood maximization. As the considered one-dimensional integral lacks an analytical form, a Monte Carlo estimator is used instead.

\subsection{Problem statements}
\label{subsubsec:problem}

For an event sequence, we consider the reconstruction of the intensity function as the main problem. To evaluate its quality, we focus on two tasks: return time prediction and event type prediction.

\textbf{Return Time Prediction}
The goal is to predict the time difference \(\Delta t_{i+1} = t_{i+1} - t_{i}\) from the current event \(i\) to the next event \(t_{i+1}\) using \(\boldsymbol{S}_{1:i}\).

\textbf{Event Type Prediction}
In our setup, events belong to \(M\) distinct types. Thus, we aim to predict the type of the next event \(\vecM_{i+1}\) using \(\boldsymbol{S}_{1:i}\).

\subsection{Continuous convolution of event sequences}
\label{subsubsec:cont_conv}

Since the time intervals \([t_{j}, t_{j+1}]\) are not necessarily equal, the considered time series are non-uniform. Standard one-dimensional CNNs are designed for equally spaced data. However, we can extend this concept to handle non-uniform cases.

The considered event sequence \(\boldsymbol{S}\) can be rewritten in functional form as follows:

\begin{equation}
\label{eq:cont_seq}
    \vecM(t) = \sum_{j=1}^{k} \vecM_{j} \delta(t - t_{j}),
\end{equation}
where \(\delta(\cdot)\) is the Dirac delta function.

Let \(\vecK(t) \colon \mathbb{R} \to \mathbb{R}^{d_{out} \times d}\) be a kernel function. To make the model causal and prevent peeking into the future, we set \(\vecK(t) = 0\) for \(t < 0\). The continuous convolution is defined as:

\begin{equation}
\label{eq:cont_conv}
    \vecY(t) = (\vecK \ast \vecM)(t) = \int_{0}^{\infty} \vecK(t - \tau) \vecM(\tau) \, d\tau = \sum_{j = 1}^{k} \vecK(t - t_{j}) \vecM_{j}.
\end{equation}

The last equality holds due to the specific form of \(\vecM(t)\) given in~\eqref{eq:cont_seq}.

In our neural network layer, the vector function of time \(\vecY(t)\) can be used in two ways. The first way involves querying \(\vecY(t)\) at a specific point in time to obtain a representation. The second way involves using \(\vecY(t)\) as input to the next convolutional layer. In this case, we limit \(\vecY(t)\) to the set \(\{t_1, \ldots, t_k\}\), obtaining a sequence of vectors \(\{\vecY_1, \ldots, \vecY_k\}\) with \(\vecY_i = \vecY(t_i)\). We then use the functional form~\eqref{eq:cont_seq} as input to the next layer.

The scheme of this continuous convolutional layer is presented in~\autoref{fig:cont_conv_layer}. This approach allows us to maintain the temporal structure across all model layers and query the embedding at any point in time without any parametric interpolations and heuristics.

\subsection{COTIC model architecture}
\label{subsubsec:our_ccnn_model}
\begin{figure}[ht]
    \centering
    \includegraphics[scale=0.7]{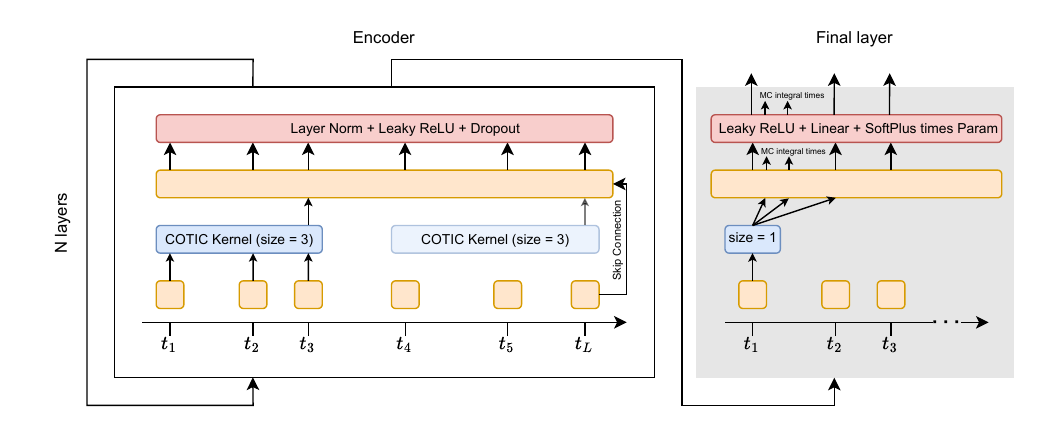}
    \usecaption{cotic}
    \label{fig:cotic_arch}
\end{figure}

Continuous one-dimensional convolution introduced by~\eqref{eq:cont_conv} allows us to get a new continuous representation of an input event sequence. The architecture can be found in \autoref{fig:cotic_arch}. In the final layer, the intensity does not use the embedding of the timestamp being predicted. This is because the model cannot know in advance the exact moment an event will occur. In other words, the intensity at a given timestamp is determined by the limit approaching from the left.

\paragraph{Kernel Parametrization}

The convolution kernel \(\vecK\) can be parameterized by any function that maps time to a matrix. However, it has been shown that a linear kernel is the optimal choice, as it facilitates model optimization.

Firstly, let's adhere to the classic kernel size constraint. While the architecture allows for an unlimited number of points in the past, as shown in \eqref{eq:cont_conv}, this would result in quadratic complexity with respect to the sequence length, negating the benefits of the convolutional approach. Therefore, we use a fixed kernel size in conjunction with a dilation mechanism.

Let's consider a straightforward approach for the final layer without any optimizations, where we additionally query \( m \) extra intensities between the timestamps. Assume we have time and feature sequences \( S = \{t_i\}_{i=1}^l \) and \( X = \{\mathbf{x}_t\}_{t=1}^l \). Here $l$ is the sequence length for a given event sequence.

For all the time intervals, we generate \( m \) randomly located points, resulting in the sequence \( S_{\text{sim}} = \{\tau_j\}_{j=1}^{(m+1)(l-1)+1} \). Here, for all \( i \in \{1, \dots, l\} \), \( j = (m+1)(i-1) + 1 \): \( \tau_j = t_i \).

Assume we have a kernel function \( \mathbf{k}(t): \mathbb{R} \to \mathbb{R}^{d_{in} \times d_{out}} \) and \( \mathbf{x} \in \mathbb{R}^{d_{in}} \). The complexity of one kernel function call will be $O(d_{in}d_{out})$.

We want to perform the convolution between the input function \( x(t) = \sum_{i=1}^{l} \mathbf{x}_i \delta(t - t_i) \) and the kernel function for each time in \( S_{\text{sim}} \). In this case, the output of the layer will be:

\begin{align*}
    y(t) &= \int_{-\infty}^{+\infty} \mathbf{k}(t - \tau)^\top x(\tau) \, d\tau = \int_{-\infty}^{+\infty} \mathbf{k}(t - \tau)^\top \sum_{i=1}^l \mathbf{x}_i \delta(\tau - t_i) \, d\tau =\\&= \sum_{i=1}^{l:t_{l+1} \geq t} \mathbf{k}(t - t_i)^\top \mathbf{x}_i \approx \sum_{i=l-\kappa}^{l:t_{l+1} \geq t} \mathbf{k}(t - t_i)^\top \mathbf{x}_i
\end{align*}

Here, we limit the number of points to a kernel size \(\kappa\). Consequently, the time complexity of this algorithm will be \(O(\kappa m l d_{in} d_{out})\), as we need to compute the intensity for \((m+1)(l-1)+1\) points. This complexity applies to both the kernel function calls and the convolution itself. Memory complexity is the same as well. Thus, we want to optimize this procedure.

To make the algorithm more time-efficient, we use a fixed set of uniform random variables \(\mathbf{u} = (u_1, \dots, u_m)^\top\) instead of generating the in-between times independently. Additionally, let's assume our kernel is linear:

\[ 
\mathbf{k}(\alpha t_1 + \beta t_2) = \alpha \mathbf{k}(t_1) + \beta \mathbf{k}(t_2); ~\alpha, \beta \in \mathbb{R} 
\]

This assumption limits the family of functions, but it allows for deeper and broader architectures due to the reduced time complexity.

Let's define the sequence \(S_{\text{sim}}\) as:

\[ 
S_{\text{sim}_j} = \begin{cases} 
t_i, & \text{if } j=(m+1)(i-1)+1 \\ 
t_i + u_s \Delta t_i, & \text{else, } s = (j - 2)\!\!\!\! \mod m + 1 \text{, } i=\lfloor(j-1)/(m+1)\rfloor 
\end{cases}
\]

This sequence includes both the original timestamps and additional points sampled uniformly within the intervals between them. The output value for each time in \(S_{\text{sim}}\) is:

\begin{align*}
    y(t_{\text{sim}_j}) &= \sum_{\nu=l-\kappa}^{l:t_{l+1} >= t_{\text{sim}_j}} \mathbf{k}(t_{\text{sim}_j} - t_{\nu})^\top \mathbf{x}_{\nu}=\\ &= \sum_{\nu=l-\kappa}^{l:t_{l+1} >= t_{\text{sim}_j}} \left( \mathbf{k}(t_i - t_{\nu})^\top \mathbf{x}_{\nu} + \Delta t_i \mathbf{k}(u_s)^\top \mathbf{x}_{\nu} \right) 
\end{align*}

This approach leverages the linearity of the kernel function, simplifying the computation. The time complexity of this method is \(O(\kappa l d_{in} d_{out} + m d_{in} d_{out} + m l d_{out})\), the time complexity of the kernel calls is \(O(\kappa l d_{in} d_{out} + m d_{in} d_{out})\).

To further optimize the algorithm, we note that if the kernel \(k(t)\) is linear and can be expressed as \(k(t) = tW\), where \(W \in \mathbb{R}^{d_{out} \times d_{in}}\), thus we can precompute \(W \mathbf{x}_i\). This transforms the convolution computation into:
\begin{align*}
    y(t_{\text{sim}_j}) &= \sum_{\nu=l-\kappa}^{l:t_{l+1} >= t_{\text{sim}_j}} (t_{\text{sim}_j} - t_{\nu})(W \mathbf{x}_{\nu}) \\&= \sum_{\nu=l-\kappa}^{l:t_{l+1} >= t_{\text{sim}_j}} (t_i - t_{\nu})(W \mathbf{x}_{\nu}) + \Delta t_i u_s (W \mathbf{x}_{\nu})
\end{align*}

In this case, the time complexity is reduced to \(O(l d_{in} d_{out} + \kappa l d_{out} + m l d_{out})\). It is worth noting, that this optimization is useful for both inner layers and the final layer.

To enhance the flexibility of our model, we introduce a bias term in the linear kernel. The kernel function now becomes:
\[ 
k(t) = t W + B, 
\]
where \(B \in \mathbb{R}^{d_{out} \times d_{in}}\) is a bias term. With this adjustment, the convolution computation is updated to:
\[ 
y(t_{\text{sim}_j}) = \sum_{|\nu : t_{\nu} < t_{\text{sim}_j}| = \kappa} (t_{\text{sim}_j} - t_{\nu})(W \mathbf{x}_{\nu}) + (B\mathbf{x_{\nu}}).
\]

\paragraph{Return Time and Event Type Prediction}

\label{sec:expect}

When using the intensity to model an event sequence, a generative model is constructed, allowing for the determination of the expected return time and event type using the definition of the intensity function. Specifically:

\begin{equation*}
    \mathbb{E}[t] = \int_{0}^{+\infty} \lambda(t)t \exp\left(-\int_{0}^{t}\lambda(\tau)d\tau\right) dt = \int_{0}^{+\infty} \exp\left(-\int_{0}^{t}\lambda(\tau)d\tau\right) dt
\end{equation*}
\begin{equation*}
    \mathbb{E}[C] = \arg \max_c \lambda_c(\mathbb{E}[t])
\end{equation*}

However, computing these integrals directly poses challenges. Instead of calculating this integral, authors typically append an additional layer and train the downstream task head. We propose an alternative approach: normalizing the data, limiting the integral, and estimating the error.

\textbf{Normalization.} Assume the data follows a Poisson process, ignoring historical dependencies, and scale the data so that the expected intensity is one. To eliminate potential outliers, consider the 99th percentile of all return times. This requires solving:

\begin{equation*}
    \mathrm{mean}(t < t_{99}) = \frac{\int_{0}^{t_{99}} \lambda t \exp(-\lambda t) \, dt} {\int_{0}^{t_{99}} \lambda \exp(-\lambda t) \, dt}
\end{equation*}

The obtained intensity serves as the scaling factor for normalization. Multiply the times by this factor.

\textbf{Approximation.} Given the normalized data, we limit the bounds for the integral:
\begin{equation*}
    \mathbb{E}[t] \approx \int_{0}^{T} \exp\left(-\int_{0}^{t}\lambda(\tau)d\tau\right) dt. 
\end{equation*}
The error estimate for this approximation is \(\exp(-T)\). This can be constrained to any desired value; in our case, we limited the error to \(10^{-6}\). The resulting integral can be computed using the trapezoidal rule with predefined steps; we used a step size of \(10^{-2}\).
\section{Experiments}\label{sec:experiments}

\subsection{Datasets}\label{subseq:datasets}

We employ a variety of event sequence datasets\footnote{Datasets can be found in \url{https://drive.google.com/drive/folders/1vxNhcbgHvq9CfW9-RhTZ2yTovFQ5F4XD}}, that are common in the field, including Retweet~\cite{zhao2015seismic}, Amazon~\cite{amazon}, StackOverflow~\cite{leskovec2014snap}, Transactions~\cite{fursov2021gradient}, Mimic-II~\cite{johnson2016mimic}, LinkedIn~\cite{xu2017dirichlet} and MemeTrack~\cite{leskovec2014snap} to benchmark our method against others.
For comprehensive statistics for these seven datasets and additional insights about each dataset, please refer to~\ref{sec:datasets_details}.

\subsection{Details of comparison}
\label{subseq:baselines}

For all experiments, we split the datasets into three parts (train, test, and validation sets) in an 8:1:1 ratio. We train the models using the Adam optimizer~\cite{kingma2014adam} for a maximum of 100 epochs.

To compare with COTIC, we consider top-performing approaches from various architectures, including TPP-based RNNs (RMTPP \cite{rmtpp_paper}, Neural Hawkes \cite{neuralhawkes_paper}, ODETPP~\cite{chen2021neural}), TPP-based transformers (THP~\cite{zuo2020transformer}, THP2SAHP, AttNHP~\cite{yang2022transformer}), and adaptations of existing CNNs for processing non-uniform event sequences (WaveNet~\cite{vanwavenet}, CCNN~\cite{shi2021continuous}). Details of these methods, along with technical details on the implementations used, are provided in~\ref{sec:appendix_baselines} and ~\ref{sec:details}.

\begin{table*}[!t]
\vspace{0.1cm}
\adjustbox{width=\textwidth}{
\begin{tabular}{cccccccccc}
\hline
Dataset & RMTPP & Neural Hawkes & ODETPP & THP & THP2SAHP & AttNHP & WaveNet & CCNN & COTIC             \\ \hline
Retweet & 
$0.030\pm0.000$  &  $\underline{0.029\pm0.001}$  &  $144.707\pm140.205$  &  $0.043\pm0.014$  &  $0.060\pm0.019$  &   $\underline{0.029\pm0.000}$   & $0.224\pm 0.021$  &  $0.138\pm0.027$  &  $\mathbf{0.0277\pm0.000}$
\\
Amazon & 
$-$  &    $89.320\pm3.320$  &  $62381\pm3338$  &  $113.470\pm4.180$  &  $98.750\pm8.400$  &   $347.344\pm0.047$  & $\underline{41.950\pm0.600}$  &  $58.900\pm0.010$  &  $\mathbf{35.514\pm0.159}$     \\
SO &
$0.839\pm0.001$  &   $0.841\pm0.001$  &  $0.843\pm0.001$  &  $13.480\pm0.840$   &  $11.980\pm0.270$  &   $0.857\pm0.020$    & $\underline{0.676\pm0.008}$  &  $1.000\pm0.070$  &  $\mathbf{0.500\pm0.004}$
\\
Transactions & $0.842\pm0.001$  &   $2.072\pm0.709$  &  $35.147\pm59.416$  &  $1.480\pm0.350$   &  $1.240\pm0.060$  &   $0.839\pm0.001$   & $\underline{0.688\pm0.010}$  &  $0.821\pm0.004$  &  $\mathbf{0.659\pm0.012}$* 
\\
Mimic & 
$\underline{0.354\pm0.001}$  &    $0.415\pm0.041$  &  $0.367\pm0.007$  &  $0.584\pm0.013$  &  $0.518\pm0.026$  &   $0.365\pm0.018$   & $3.280\pm1.140$  &  $0.555\pm0.022$  &  $\mathbf{0.249\pm0.011}$
\\
LinkedIn & $2.521\pm0.001$  &   $2.483\pm0.015$  &  $6.057\pm4.371$  &  $9.890\pm0.290$   &  $9.520\pm0.220$  &   $2.469\pm0.002$   & $\mathbf{1.490\pm0.020}$  &  $2.560\pm0.350$  &  $\underline{1.541\pm0.022}$
\\
MemeTrack & 
$-$  &    $\underline{46.803\pm0.009}$  &  $59.759\pm12.409$  &   $97.770\pm0.650$   &   $99.690\pm8.690$  &   $\mathbf{46.359\pm0.154}$  & $63.080\pm1.460$  &   $138.470\pm9.180$  &  $48.370\pm0.054$
\\[0.3mm] \hline
\# of wins &
$0$ &
$0$ & 
$0$ & 
$0$ & $0$ & $\underline{1}$ & 
$\underline{1}$ &  $0$ & $\mathbf{5}$ \\
\bottomhline
\end{tabular}
 }
 \caption{Return time prediction MAE $\downarrow$ results. The best result for each dataset is in bold; the second-best model result is underlined. The asterisk indicates that for this dataset we used linear regression model on top of the embeddings. Better to view in zoom.}
 \label{tab:return_time_results}
\end{table*}

\begin{table*}[!t]
\vspace{0.1cm}
\resizebox{\textwidth}{!}{%
\begin{tabular}{cccccccccc}
\hline
Dataset & RMTPP & Neural Hawkes & ODETPP & THP & THP2SAHP & AttNHP & WaveNet & CCNN & COTIC
\\ \hline
Retweet & 
$0.555\pm0.010$  &   $0.605\pm0.003$  &  $0.525\pm0.025$  &  $0.499\pm0.031$  &  $0.518\pm0.022$  &  $0.578\pm0.004$   & $\underline{0.606\pm0.001}$  &  $0.349\pm0.043$  &  $\mathbf{0.608\pm0.001}$
\\
Amazon & 
$-$  &   $\mathbf{0.872\pm0.009}$  &  $0.041\pm0.008$   &  $0.331\pm0.005$  &  $0.331\pm0.006$  &  $0.189\pm0.082$   & $0.564\pm0.005$  &  $0.095\pm0.021$  &  $\underline{0.614\pm0.002}$
\\
SO & 
$0.366\pm0.117$   &  $0.431\pm0.017$  &  $0.444\pm0.005$  &  $0.432\pm0.001$   &  $0.432\pm0.001$  &  $0.327\pm0.063$  & $\underline{0.470\pm0.000}$  &  $0.056\pm0.010$  &  $\mathbf{0.592\pm0.010}$
\\
Transactions & 
$0.252\pm0.109$   &  $0.293\pm0.026$  &  $0.305\pm0.018$  &  $0.315\pm0.000$   &  $0.315\pm0.000$  &  $0.189\pm0.082$   & $\mathbf{0.369\pm0.011}$  &  $0.017\pm0.003$  &  $\underline{0.337\pm0.004}$*
\\
Mimic & 
$0.874\pm0.034$   &   $\mathbf{0.930\pm0.012}$  & $0.855\pm0.029$  &  $0.699\pm0.176$  &  $0.788\pm0.067$  &  $0.893\pm0.007$   & $0.300\pm0.250$  &  $0.013\pm0.008$  &  $\underline{0.911\pm0.009}$
\\
LinkedIn &
$0.148\pm0.002$   &   $0.255\pm0.004$  &  $0.210\pm0.010$  &  $0.132\pm0.009$  &  $0.147\pm0.008$  &  $\mathbf{0.278\pm0.008}$   & $0.262\pm0.011$  & $0.007\pm0.001$  &  $\underline{0.277\pm0.010}$
\\
MemeTrack & 
$-$    &   $\underline{0.117\pm0.037}$  &  $0.010\pm0.008$  &  $0.013\pm0.002$  &  $0.014\pm0.001$  &  $0.051\pm0.009$  & $0.050\pm0.004$  & $0.005\pm0.001$  &  $\mathbf{0.124\pm0.001}$ 
\\[0.2mm] \hline
\# of wins & 
$0$ & $\underline{2}$ & $0$ & $0$ & $1$ & $1$ & $1$ & $0$ & $\mathbf{3}$ \\ 
\bottomhline
\end{tabular}
}
\caption{Event type prediction accuracy $\uparrow$ results. The best result for each dataset is in bold; the second-best model result is underlined. The asterisk indicates that for this dataset we used logistic regression model on top of the embeddings. Better to view in zoom.}
\label{tab:event_type_results}
\end{table*}

\subsection{Main experiments} \label{subsubsec:main_experiments_results}

\textbf{Model Meta-Parameters}
A key question for researchers is which meta-parameters to use to achieve the best possible performance. Our goal was to develop a model that is universal and performs well with default parameters or minimal tweaking. For baseline models, we used the parameters suggested by their respective authors. For our model, we used a 9-layer architecture with a kernel size of 3 and a hidden size of 512 to balance the limitations of the linear kernel. Inner experiments with Optuna optimization~\cite{akiba2019optuna} indicated that this setup is the best or close to the best across several datasets. The return time and event type were computed using the expectations formula provided in~\autoref{sec:expect}. However, there were two datasets where we deviated from this setup:
\begin{itemize}
    \item \textbf{MemeTrack}: The dataset's high variety of event types combined with a limited amount of data—approximately 300,000 events in total—made it challenging to model. Reducing the number of layers to 2 significantly improved performance, likely by preventing overfitting and allowing the model to generalize better from the limited data available.
    \item \textbf{Transactions}: Our assumptions for return time inference hold only if the data is in continuous time. However, the timestamps in this dataset are integers. Although the model can capture these dependencies, our error estimations are incorrect. Therefore, instead of computing expectations, we trained Linear Regression and Logistic Regression models on top of the embeddings. Also, to reduce the complexity of the Linear Regression and Logistic Regression training, we reduced the hidden size to 64.
\end{itemize}

\textbf{Return Time and Event Type Prediction.} We compared the COTIC model with baseline models on two commonly used tasks to evaluate the quality of event sequence models: return time prediction and event type prediction. It is important to note that likelihood computation can vary between methods, making these values unsuitable for direct comparison across different models. For all experiments, we average the results and estimate standard deviations using five independent runs.

\emph{Return time prediction.}
\autoref{tab:return_time_results} presents obtained results.
As one can see, our method performs best on five out of seven datasets, demonstrating COTIC’s superiority, robustness, and consistency in return time prediction across various datasets. Additionally, there is no close competitor that consistently matches our method's performance. While WaveNet shows promising results on some datasets, it fails on the Retweet and Mimic datasets.

\emph{Event type prediction.}
Our findings regarding the event type prediction experiment are presented in~\autoref{tab:event_type_results}. 
The COTIC model exhibits high performance, being competitive against other contenders. It is either the best or second-best model, achieving the top performance on three datasets. Moreover, there is no single close competitor; when COTIC is not the best, the best model varies. This reflects COTIC's enhanced predictive abilities and consistency across various datasets.

\textbf{Downstream Problem}

Neural TPPs are generative models, and we expect them to have strong representation capabilities. Consequently, we can reuse the obtained representations for other downstream problems. To validate this, we conducted an additional experiment.

For the Transactions dataset, we predicted one of four age bins using embeddings obtained from each model. We trained a Logistic Regression classifier \cite{peng2002introduction} on top of the embeddings, performing linear probing with a frozen model~\cite{li2021efficient}. We compared COTIC with a 64-dimensional embedding to other intensity-based models and the MiniRocket model with 256 kernels~\cite{minirocket}, designed for uniform time series classification. For MiniRocket, time as a feature and one-hot encoded event types were used as inputs.

The results are presented in \autoref{tab:downstream}. Our model showed the best performance among TPP-based embeddings, while MiniRocket was slightly inferior to COTIC. We hypothesize that other TPP-based approaches provide very local representations of an event sequence at each point, whereas our convolutional layers offer greater representative power.

\begin{table}[!b]
\centering
\resizebox{\textwidth}{!}{\begin{tabular}{cccccccc}
\hline
Method             & Base & Neural Hawkes & THP & ODETPP & ATTNHP & MiniRocket & COTIC (ours) \\ \hline
Accuracy & 0.25 & 0.254 & 0.290 & 0.265 & 0.258 & \underline{0.452} & \textbf{0.459} \\
\hline
\end{tabular}}
\caption{Age bin prediction accuracy for Transactions dataset}
\label{tab:downstream}
\end{table}

\subsection{Sensitivity study}
\label{subsec:ablation_study}

\begin{figure*}[!t]
    \centering
    \includegraphics[width=.26\textwidth]{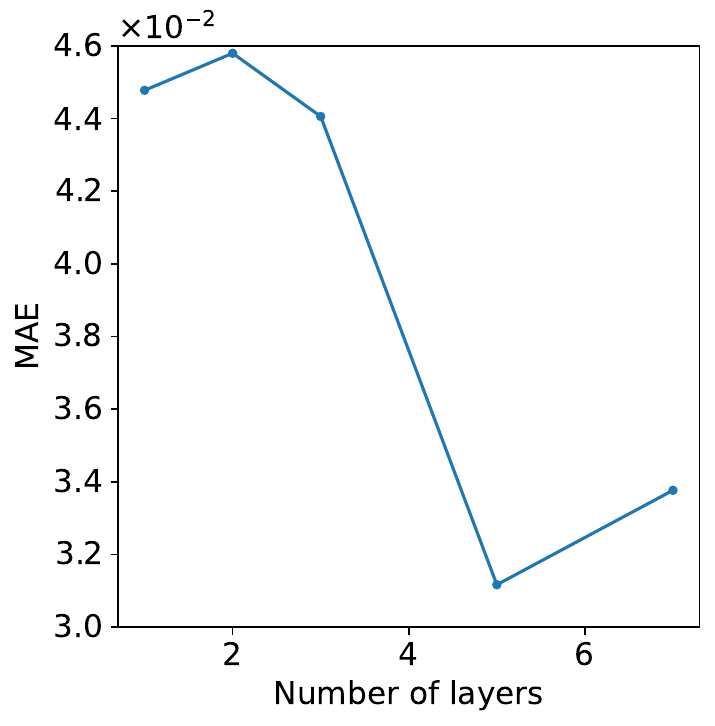}
    \includegraphics[width=.26\textwidth]{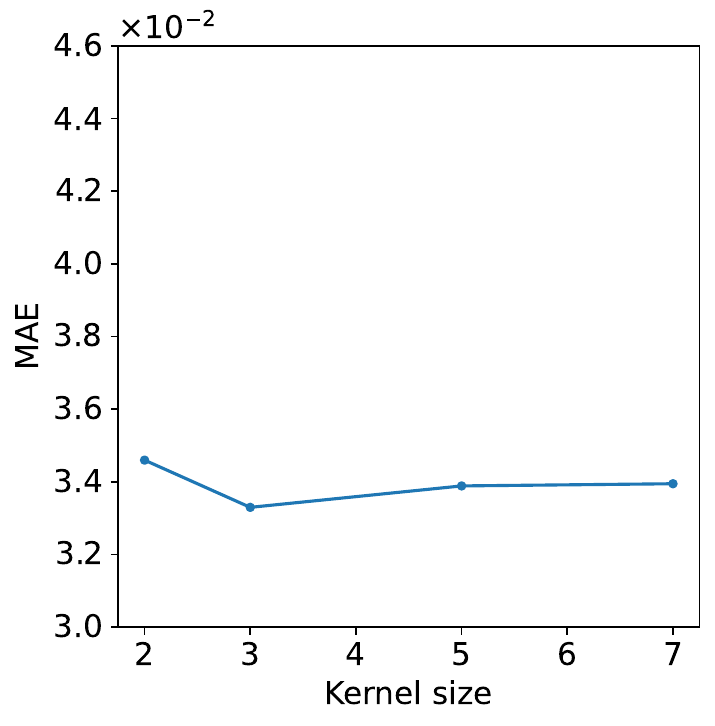}
    \includegraphics[width=.26\textwidth]{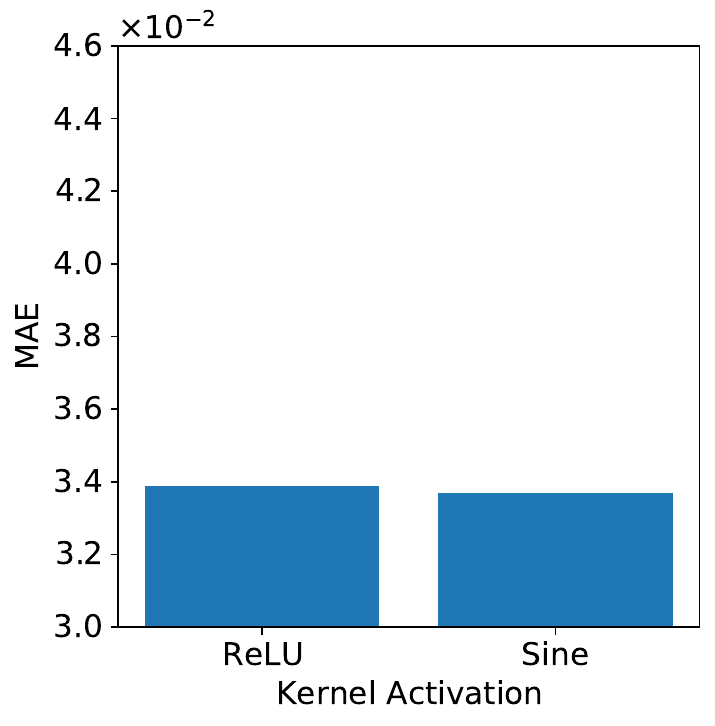}
    \usecaption{sense}
    \label{fig:abl}
\end{figure*}

Our previous claims highlighted that a high number of convolutional layers for COTIC is vital for high performance.
We also want to verify what kernel size and activation function we need.
These experiments were run on the Retweet dataset~\cite{zhao2015seismic}; the results are shown in~\autoref{fig:abl}.

There is a tangible trade-off on the number of layers: if the number of layers is too low, then the model's complexity and the receptive field size are not enough to catch data intricacies; if we make the model too deep, it becomes harder to train it, ending up with a slightly underperforming model.
However, there is little dependence on the kernel size and the activation function. 
The first result means that, given enough layers, the model already has a large enough receptive field and complexity to catch all the temporal dependencies within the dataset, and there is no need to make the model even more complex by increasing the kernel size. 
Despite that, one should be careful in interpolating this fact to all the other datasets as soon it can be the property of particular retweet sequences.
The latter experiment was to substitute the ReLU in the kernel with a new Sine activation~\cite{NEURIPS2020_53c04118}.
As one can notice, there was almost no influence on the performance.

\subsection{Synthetic Data Analysis} 
Research that revolves around sequences of events is characterized by its intensity. To facilitate a comprehensive understanding, we created synthetic event sequences sampled from the Hawkes process with known parameters, allowing us to adjust the intensity as desired. 

\begin{figure*}[!t]
    \centering 
    \includegraphics[width=\textwidth]{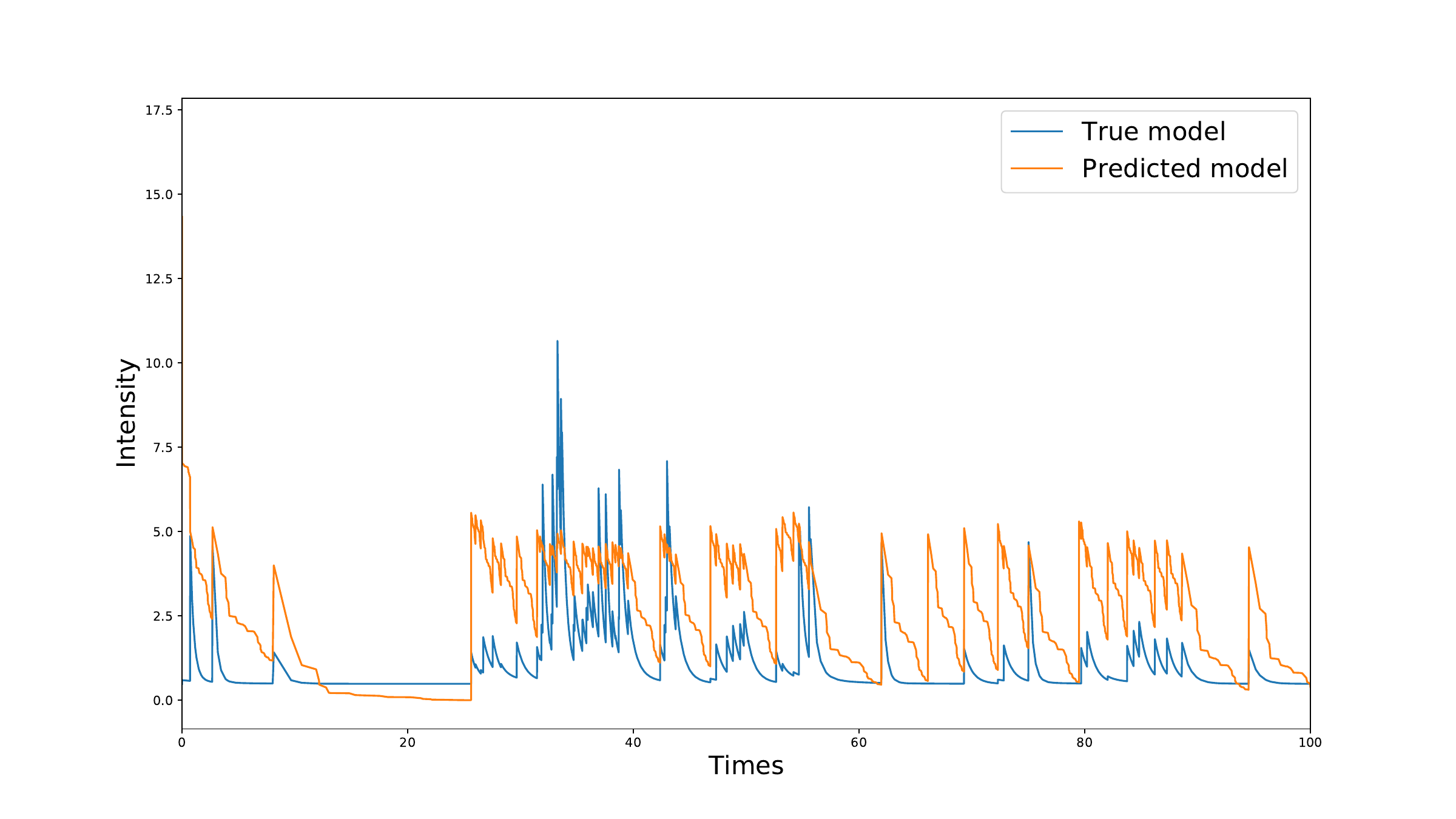}
    \usecaption{intens}
    \label{fig:intens}
\end{figure*}

In~\autoref{fig:intens}, we present a comparison between our baseline intensity and the performance of the COTIC model on the synthetic sample. The results demonstrate that COTIC closely mirrors the true distribution, accurately capturing the jump moments akin to the original Hawkes process. Furthermore, we have successfully preserved the dynamics of the series, characterized by exponential declines between events.

\section{Discussion}
On average, as indicated by the ranks in \autoref{tab:ranks}, the COTIC model demonstrates the best overall performance, followed by Neural Hawkes, WaveNet, and Attentive Neural Hawkes. However, the training time for Neural Hawkes and Attentive Neural Hawkes is prohibitive for many datasets. WaveNet is very fast, however it is not consistent as soon it does not use the Temporal Point Process framework and may fail to catch the dependencies. 
CCNN~\cite{shi2021continuous}, the closest solution to ours, underperforms in both event type and return time prediction because it was designed for different problems and considers simpler architecture. 
Therefore, the careful introduction of multiple non-uniform convolutional layers is crucial for effectively modeling event sequences using CNNs --- and COTIC excels in this task.

\section{Conclusion}

Modeling temporal point processes is a long-standing challenge, and our research addresses this using a novel approach. We leveraged a CNN to model temporal point processes, resulting in the development of our COntinuous-TIme Convolutional (COTIC) model for event sequences.

By introducing a continuous convolution function for non-uniform sequences, our approach offers several inherent advantages. These include the utilization of dilation for long-term memory, separate convolutions for various event types, and variable depth allowing for smooth time transformations, which is essential for processing non-uniform event sequences.

The experimental results highlight the quality and efficiency of the COTIC model. It outperforms both recurrent and transformer-based methods on datasets commonly used to benchmark temporal point process models. Additionally, COTIC offers greater efficiency in training time compared to methods with similar performance.

Given its robust performance and efficiency, the COTIC model has the potential to set a new benchmark in temporal point process modeling --- as well as exhibit numerous applications in fields that utilize event sequences.

\section{Acknowledgments}
The work was supported by the Analytical Center for the Government of the Russian Federation (subsidy agreement 000000D730321P5Q0002, Grant No.70-2021-00145 02.11.2021).

\section{Declaration of generative AI and AI-assisted technologies in the writing process}
During the preparation of this work the authors used ChatGPT in order to check the grammar and improve the word choice in some sentence. After using this tool/service, the authors reviewed and edited the content as needed and take full responsibility for the content of the published article.

\bibliographystyle{elsarticle-num} 
\bibliography{biblio}

\newpage

\section{Glossary}
\begin{enumerate}
    \item Recurrent Neural Network - an artificial neural network where connections between nodes can form a cycle.
    \item Transformer Neural Network - an artificial neural network based on the attention mechanism.
    \item Attention mechanism - a technique to selectively focus on relevant information in the data.
    \item Convolution Neural Network - a feed-forward neural network that learns features by filter optimization.
    \item Temporal Point Process - a random process composed of a time-series of discrete events that occur in continuous time.
    \item Intensity function - the expectation of the counting function increase in the infinitely small period of time.
    \item Likelihood - the probability of the model output given model parameters.
    \item Dirac delta function - a function that is zero everywhere except at zero and whose integral over the entire real line is equal to one.
    \item Algorithmic complexity - a measure of algorithmic running time given an input size.
    \item Mean absolute error (MAE) - an averaged absolute value difference between the predicted and expected output.
    \item Accuracy - a measure of the correct guesses for the classification problem.
\end{enumerate}

\newpage

\appendix

\section{Details on datasets}
\label{sec:datasets_details}

Comprehensive statistics for these seven datasets are presented in~\autoref{tab:datasets_statistics}. 

\begin{table}[!t]
\centering
\vspace{0.1cm}
\begin{tabular}{llll}
\hline
Dataset & \# of event types & Mean sequence length & \# of sequences \\
\hline
Retweet  & 3 & 108.8 & 29000\\
Amazon   & 8 & 56.5 & 7523\\
SO       & 22 & 72.4 & 33165 \\
Transactions      & 60 & 862.4 & 30000\\
Mimic    & 65 & 8.6 & 285 \\
LinkedIn & 82 & 3.1 & 2439\\
MemeTrack     & 4977 & 10.8 & 29696 \\
\bottomhline
\end{tabular}
\caption{Statistics of the used datasets}
\label{tab:datasets_statistics}
\end{table}

\paragraph{Retweets}~\cite{zhao2015seismic}: The dataset presents sequences of tweets containing an origin tweet and some follow-up tweets. We record the time and the user tag of each tweet. Further, users are grouped into three categories based on the number of their followers: "small", "medium", and "large". 
These groups correspond to the three types of events.
    
\paragraph{Amazon}~\cite{amazon}: 
    This dataset consists of around 7.5k user sequences with product reviews and product categories. All products are divided into 8 distinct groups.
    
\paragraph{StackOverflow}~\cite{leskovec2014snap}: 
    StackOverflow (SO) is a question-answering website. It rewards users with badges to promote engagement in the community, and the same badge can be given multiple times to the same user. We collect data for a two-year period and treat each user's reward history as a sequence. Each event in the sequence signifies a receipt of a particular medal.
    

\paragraph{Transactions}~\cite{fursov2021gradient}: 
    The dataset contains sequences of transaction records stemming from clients of financial institutions. For each client, we have a sequence of transactions. We describe each transaction with a discrete code, identifying the type of transaction, and the Merchant Category Code, such as a "bookstore", "ATM", "drugstore", etc. 
    
\paragraph{MIMIC-II (Electrical Medical Records)}~\cite{johnson2016mimic}: MIMIC collects patients' visits to  a hospital's ICU in a seven-year period. We treat the visits of each patient as a separate sequence where each element in the sequence contains a time stamp and a diagnosis.

\paragraph{LinkedIn}~\cite{xu2017dirichlet}: This dataset provides information on users' sequences of employment companies. There are 82 different companies comprising different event types with 2439 users in total. 

\paragraph{MemeTrack}~\cite{leskovec2014snap}: It collects mentions of 42k different memes spanning ten months, combining data from over 5000 websites with over 1.5 million documents including blog posts, web articles, etc. Each sequence in this dataset is the life-cycle of a particular meme where each event (meme occurrence) is associated with a timestamp and a website ID, serving as the event type.

\section{Baseline methods}
\label{sec:appendix_baselines}

\paragraph{RNN-based models.}
\textbf{Neural Hawkes Process}~\cite{neuralhawkes_paper}
is one of the first neural models that were proposed for event sequences. This model uses LSTM-like architecture with additional evolution of the hidden states between events.
The main benefit of this model is that it has almost no assumptions on the parametric structure for the intensity interpolation. However, Neural Hawkes is very slow in training.
\textbf{Recurrent Marked Temporal Point Processes}
RMTPP~\cite{rmtpp_paper} is another RNN-based model proposed for event sequence forecasting. In contrast to Neural Hawkes, this model assumes a parametric structure for the intensity interpolation, which makes it less expressive.
Finally, \textbf{ODE Temporal Point Process} provides a simplified version from Neural Spatio-Temporal Point Processes~\cite{chen2021neural}, which exploited a different class of parametrization for spatio-temporal point processes. In a nutshell, we model Temporal Point Process with Neural ODE state evolution via an RNN.

\paragraph{Attention-based models.}
\textbf{Attentive Neural Hawkes Process}~\cite{yang2022transformer} further develops the idea from the Neural Hawkes Process paper, but instead of an LSTM, they employed Transformer. The generative model, AttNHP, is computationally expensive, as it requires obtaining deep embedding for a possible event type to calculate its intensity at a given timestamp, and during training, likelihood calculation requires computing the intensities of many possible events.
\textbf{Transformer Hawkes Model}
Transformer is a superior architecture for NLP problems. This fact inspired the authors of ~\cite{zuo2020transformer} to use transformers in an event sequence set-up. The main benefit of Transformer compared to RNN-like models is that it does not struggle from small memory and vanishing and exploding gradients. However, the solution has some parametric assumptions on the intensity interpolation.

\paragraph{CNN-based models.}
\textbf{WaveNet} is a well-known architecture proposed in \cite{vanwavenet} designed for audio generation. 
It is one of the prominent time series processing models that adopts the ideas of convolutional neural networks.
Efficient implementation and dilated convolutions leading to a large receptive field are among the advantages of WaveNet. 
Due to these nice properties, we use this convolutional architecture as a baseline. 
As WaveNet is not directly applicable to event sequence forecasting with non-uniform times between events, we use time lag as an additional input feature.
\textbf{Continuous CNN for Nonuniform Time Series}
In a recent work~\cite{shi2021continuous}, it was proposed to use Continuous Convolutional Neural Networks (CCNNs) to model non-uniform time series. 
Although this method is the closest to the solution we present, it has notable differences and drawbacks regarding event sequence modeling.
First, the paper's authors address several problems, i.e. auto-regressive time series forecasting, speech interpolation, and intensity estimation for temporal point processes. However, the architecture itself was designed for signal restoration. In this case, there exists an actual underlying signal that can be defined at any point in time. 
For this reason, the authors propose to use only one continuous convolutional layer and allocate points evenly after that, proceeding with the standard discrete convolutions.
Therefore, the non-uniformity of time exists only in the first layer which does not consider the specifics of the event sequences modeling task.
Second, the intensity function is explicitly defined parametrically, strictly limiting the model's expressiveness.

\subsection{Technical details for baseline methods}
\label{sec:details}

In general, we used hyperparameters provided by the authors of the methods while adjusting the learning rate for some datasets to improve the performance. Further details are given below. 

\paragraph{RNN-based models.} We use two acknowledged variations of RNN-based models: Neural Hawkes and RMTPP (Recurrent Marked Temporal Point Process).
For the \emph{MemeTrack} and \emph{Amazon} datasets, RMTPP time loss quickly explodes within a single epoch, leading to the absence of results, which is reflected in our tables. For \textbf{Neural Hawkes}, \textbf{RMTPP (Recurrent Marked Temporal Point Process)}  we used PyTorch implementation from the following GitHub repo.\footnote{\url{https://github.com/ant-research/EasyTemporalPointProcess}} For a \textbf{Neural Spatio-Temporal Point Processes (ODETPP}, which combines RNN and Neural ODE, we have 
borrowed code from the same repository.

\paragraph{CNN-based models.} We also adopt two variants of one-dimensional convolutional architectures to process event sequences. 
For \textbf{WaveNet}, we used the following implementation\footnote{\url{https://github.com/litanli/wavenet-time-series-forecasting}}.
We note that \textbf{WaveNet} does not take into the non-uniformity of a sequence, but it is a strong baseline that takes the roots in CNN ideas.
For \textbf{CCNN}, following the original architecture\footnote{\url{https://github.com/shihui2010/continuous_cnn}}, we use a model consisting of one continuous convolutional layer followed by standard discrete convolutions and a particular output layer reconstructing intensity in the exponential form given in~\cite{shi2021continuous}.
Consequently, the likelihood function becomes a double exponent, leading to a robust numerical instability caused by overflows in exponent.
To prevent this, we normalize the datasets and use strict early-stopping conditions while training for CCNN. 

\paragraph{Attention-based models.} Finally, we use three SOTA variants of attention-based models specifically tailored to TPP.
\textbf{THP (Transformer Hawkes Process)} has the implementation from the repo\footnote{\url{https://github.com/SimiaoZuo/Transformer-Hawkes-Process}}. Subsequently, the model was introduced into the pipeline, in which COTIC training took place.
Moreover, we corrected the intensity function and, consequently the likelihood, by excluding the current hidden state to make the model consistent.
\textbf{THP2SAHP} combines THP~\cite{zuo2020transformer} and the calculation of the intensity function from~\cite{zhang2020self} implemented in the repo\footnote{\url{https://github.com/QiangAIResearcher/sahp_repo}}.
Since the intensity (hence, the likelihood) calculation in this approach uses the current hidden state, it cannot be compared with other architectures w.r.t. the likelihood. Finally, we employed realization of \textbf{Attentive Neural Hawkes Process} from previously mentioned Ant Research repository.


\end{document}